\documentclass[conference]{IEEEtran}
\IEEEoverridecommandlockouts
\usepackage{cite}
\usepackage{amsmath,amssymb,amsfonts}
\usepackage{algorithmic}
\usepackage{graphicx}
\usepackage{textcomp}
\usepackage{xcolor}
\usepackage{hyperref}
\usepackage{multirow}
\usepackage{algorithm}
\usepackage{algorithmic}
\def\BibTeX{{\rm B\kern-.05em{\sc i\kern-.025em b}\kern-.08em
    T\kern-.1667em\lower.7ex\hbox{E}\kern-.125emX}}
\begin{document}

\title{Connection Reduction of DenseNet for Image Recognition\\
}

\author{\IEEEauthorblockN{Rui-Yang Ju, Jen-Shiun Chiang, Chih-Chia Chen, and Yu-Shian Lin}
\IEEEauthorblockA{Department of Electrical and Computer Engineering, Tamkang University}
\IEEEauthorblockA{Yingzhuan Rd., Tamsui Dist., New Taipei City, 251301, Taiwan}
\IEEEauthorblockA{\{jryjry1094791442, jsken.chiang, crystal88irene, abcpp12383\}@gmail.com}
\thanks{Corresponding author: Jen-Shiun Chiang (email: jsken.chiang@gmail.com).}}

\maketitle

\begin{abstract}
Convolutional Neural Networks increase depth by stacking convolution layers, and deeper network models perform better in image recognition. Empirical research shows that simply stacking convolution layers does not make the network train better, and skip connection (residual learning) can improve network model performance. For the image classification tasks, models with global densely connected architectures perform well in large datasets like ImageNet, but they are not suitable for small datasets such as CIFAR-10 and SVHN. Different from dense connections, we propose two new algorithms to connect layers in this paper. $Baseline$ is a densely connected network, and the networks connected by the two new algorithms are named $ShortNet_1$ and $ShortNet_2$, respectively. The experimental results of image classification on CIFAR-10 and SVHN show that $ShortNet_1$ has a 5\% lower test error rate and 25\% faster inference time than $Baseline$. $ShortNet_2$ speeds up inference time by 40\% with less loss in test accuracy. Code and pre-trained models are available at \url{https://github.com/RuiyangJu/Connection_Reduction/}
\end{abstract}


\section{Introduction}
Convolutional Neural Networks (CNN) have become the main network architecture in Computer Vision tasks. From the evaluation results of various network models in image classification, the deeper network models always have higher test accuracy. Therefore, researchers began to design the neural network architecture by stacking more convolution layers. However, the researchers found that a bottleneck of increasing depth occurs when simply stacking convolutional layers to a certain extent. ResNet \cite{he2016deep} first proposes residual learning to solve this bottleneck, and the network can train deeper models to achieve higher accuracy. On this basis, DenseNet \cite{huang2017densely} inherits and improves the concept of skip connection, making all layers connect to each other, which obtains state-of-the-art performance in image classification.

Experiments \cite{zhu2018sparsely} show that not all connections between layers are positive, and the large memory access and slow inference time of the model have become shortcomings of DenseNet. We note that DenseNet cannot be widely used in prediction tasks such as semantic segmentation and object tracking because of the above shortcomings. Therefore we propose two new algorithms for connecting between layers to replace dense connection in this paper.

This research work has two main contributions: The applying conditions of the newly proposed algorithms are not difficult and they can completely replace the dense connection. These two new algorithms can be applied to DenseNet deformation network in the future and have more applications. And this work also proves that not all connections in DenseNet play positive roles for small datasets, and appropriately reducing the connections between layers can improve the efficiency of the network model.

\section{Related Works}
In order to solve the gradient vanishing problem, ResNet proposes skip connection, which expresses the output of the convolution layer as a linear superposition of the input and a nonlinear transformation of the input. The key advantage of skip connection is that the distance between feature layers is short during backpropagation. 

However, the extreme connection method of DenseNet makes it high computing costs in the application process. Microsoft Research proposed Log-DenseNet \cite{hu2017log}, which designs a fixed number of skip connections to reduce the total number of connections from $L^2$ to $Llog_2L$. In object recognition and semantic segmentation tasks, Log-DenseNet outperforms DenseNet, and the speed is improved from $L^2$ to $Llog_2L$. Liu \emph{et al.} \cite{liu2018sparsenet} proposed another approach (SparseNet) to sparse DenseNet, reducing the number of connections from $L^2$ to $L$ while increasing the width of the neural network. The inference rate of SparseNet is 2.6 times faster than DenseNet in image classification. Chao \emph{et al.} \cite{chao2019hardnet} analyzed that the memory traffic generated by extracting the feature map of the middle part is a factor affecting the inference speed of the network model, and proposed HarDNet. Therefore, HarDNet changes the global dense connection to the harmonic dense connection, and reduces the loss of accuracy by increasing the output weight, which achieves high efficiency in terms of low memory traffic. On this basis, ThreshNet \cite{ju2022threshnet} proposed threshold mechanism, which is used to judge the choice of dense connection and harmonic dense connection, and reset the number of channels in each block. In addition, there are researches \cite{wang2021sparsenet}\cite{ju2021new} to improve the Blocks in DenseNet. The former screen Blocks by maintaining an array recording weight, and the latter completes the network architecture by combining Blocks with two different connection methods.

\begin{figure}[htbp]
\centerline{\includegraphics[width=\linewidth]{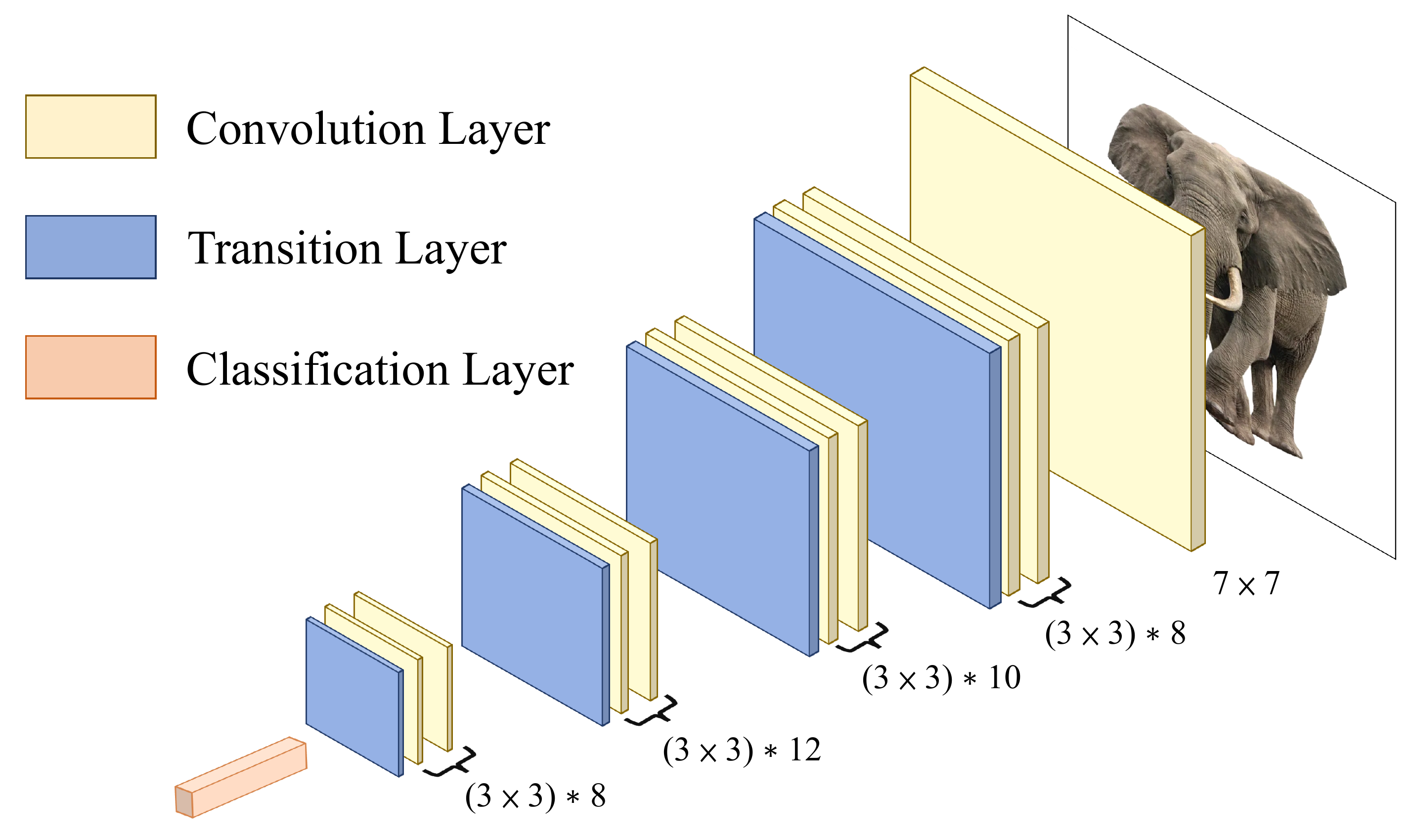}}
\caption{Neural network with simple stacked convolutional layers.}
\label{fig1}
\end{figure}

\section{Research Method}
\subsection{Baseline}
In order to fairly compare the dense connection and the connections using the two algorithmic approaches, we do not  use any tricks for the design of the neural network architecture.  The network architecture only consists of convolution layers and transition layers. Fig. \ref{fig1} represents a neural network with 43 layers, including 39 convolution layers and 4 transition layers. We design two network models with different depths, which are 43 layers and 53 layers, respectively, as shown in TABLE \ref{tab1}.

The feature maps of each layer are the same size and can be connected in the channel dimension. The nonlinear combination function $H\left(\right)$ in the Block of $Baseline$ adopts the structure of $BN + ReLU + 3 \times 3 \ Conv$. Each layer of convolution in all blocks outputs $k$ feature maps, which means the number of channels of the obtained feature map is $k$. $k$ is called growth rate, which is a hyperparameter. Referring to DenseNet \cite{huang2017densely} setting for growth rate, we set $k$ to 32 in image classification task, which can achieve higher performance.

In addition, because the input of the latter layer would be very large, DenseNet uses the bottleneck layer to reduce the amount of calculation. Bottleneck adopts the architecture of $BN + ReLU + 1 \times 1 \ Conv + BN + ReLU + 3 \times 3 \ Conv$. The purpose of this paper is to compare the efficiency of different ways of connecting, so the bottleneck is not used to optimize the network model in our experiment.

The connection between the layers in $Baseline$ network adopts the dense connection of DenseNet. Each layer in  $Baseline$ network will be connected with all the previous layers on the channel. For the network with $n$ layers, there are $\frac{n(n+1)}{2}$ connections in total, and the formula is as follows:

\begin{equation}
x_n=\ H_n\left(\left[x_1,\ \cdots,\ x_{n-1}\right]\right)
\label{eq1}
\end{equation}

As shown in Fig. \ref{fig1}, we design a transition layer between two adjacent Blocks. The structure of the transition layer is $BN + ReLU + 1 \times 1 \ Conv + 2 \times 2 \ AvgPool$, and its effect is to reduce the size of the feature map. The transition layer can generate $\theta\times output$ channel features by convolution, where $\theta\in(0,1]$ is the compression rate. When $\theta=1$, the number of features does not change through the transition layer. We use $\theta=0.5$ for $Baseline$, which means the number of features is reduced by half.

\begin{table}[htbp]
\caption{Baseline Network Model Architecture}
\centering
\setlength{\tabcolsep}{5.5mm}{
\begin{tabular}{|c|c|c|}
\hline
\textbf{Model} & \textbf{Baseline-43} & \textbf{Baseline-53} \\ \hline
Convolution & \multicolumn{2}{|c|}{7 × 7 Conv} \\ \hline
\multirow{3}{*}{\centering Block-1} & (3 × 3 Conv) * 8 & (3 × 3 Conv) * 8 \\ \cline{2-3}
 & \begin{tabular}[c]{@{}c@{}}1 × 1 Conv \\ 2 × 2 AvgPool\end{tabular} & \begin{tabular}[c]{@{}c@{}}1 × 1 Conv \\ 2 × 2 AvgPool\end{tabular} \\ \hline
\multirow{3}{*}{\centering Block-2} & (3 × 3 Conv) * 10 & (3 × 3 Conv) * 12 \\ \cline{2-3}
 & \begin{tabular}[c]{@{}c@{}}1 × 1 Conv \\ 2 × 2 AvgPool\end{tabular} & \begin{tabular}[c]{@{}c@{}}1 × 1 Conv \\ 2 × 2 AvgPool\end{tabular} \\ \hline
\multirow{3}{*}{\centering Block-3} & (3 × 3 Conv) * 12 & (3 × 3 Conv) * 16 \\ \cline{2-3}
 & \begin{tabular}[c]{@{}c@{}}1 × 1 Conv \\ 2 × 2 AvgPool\end{tabular} & \begin{tabular}[c]{@{}c@{}}1 × 1 Conv \\ 2 × 2 AvgPool\end{tabular} \\ \hline
\multirow{3}{*}{\centering Block-4} & (3 × 3 Conv) * 8 & (3 × 3 Conv) * 8 \\ \cline{2-3}
 & \begin{tabular}[c]{@{}c@{}}1 × 1 Conv \\ 2 × 2 AvgPool\end{tabular} & \begin{tabular}[c]{@{}c@{}}1 × 1 Conv \\ 2 × 2 AvgPool\end{tabular} \\ \hline
\multicolumn{3}{|c|}{Linear Classification} \\ \hline
\end{tabular}
}
\label{tab1}
\end{table}

\begin{figure}[htbp]
\centerline{\includegraphics[width=\linewidth]{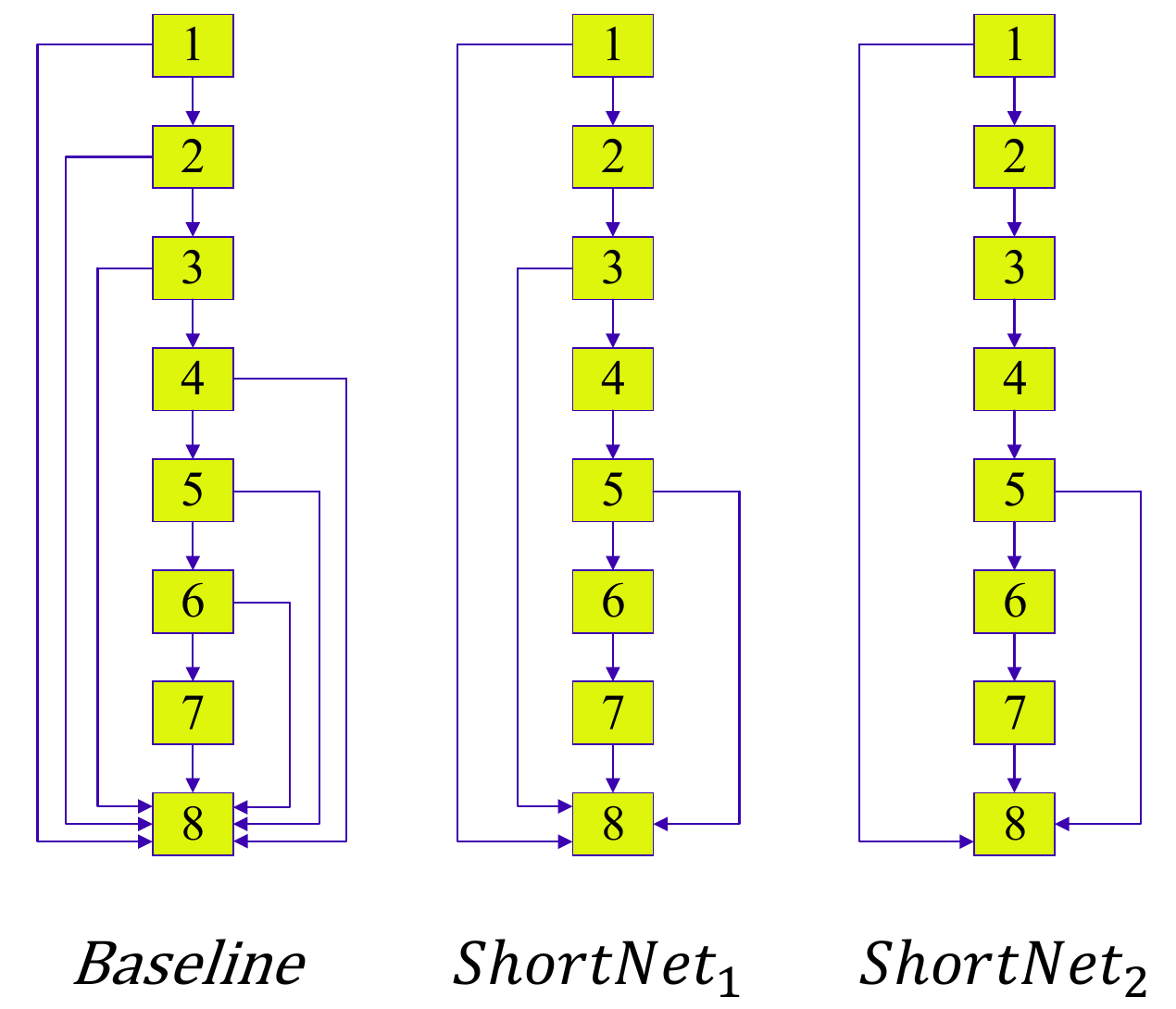}}
\caption{Example of the connections between the $8th$ convolutional layer and the previous layers.}
\label{fig2}
\end{figure}

\begin{algorithm}
\caption{$Shortnet_1$}
\label{alg_1}
\begin{algorithmic}
\REQUIRE {Stack 3 × 3 Convolution Layer in each block}
\ENSURE {$BN + ReLU + 3 \times 3 \ Conv$}
\FOR{layer $n$ is odd}
\STATE {layer $n$ connect to layer $2^0$}
\FOR{$i$ in $2^1$ to $2^5$}
\STATE {let $n$ connect to layer $i$ ($i \leq n$ and $i$ is even)}
\ENDFOR
\ENDFOR
\FOR{layer $n$ is even}
\STATE {layer $n$ connect to layer $2^0$}
\FOR{$i$ in $2^1$ to $2^5$}
\STATE {let $n$ connect to layer $i$ ($i \leq n$ and $i$ is odd)}
\ENDFOR
\ENDFOR
\end{algorithmic}
\end{algorithm}

\subsection{$ShortNet_1$}
The first new connection method we propose is different from the connection in SparseNet \cite{liu2018sparsenet} that skips the middle part and only connects the farthest layer and the nearest layer. Instead of deleting connections from a specific concentrated part, we perform connection reductions at intervals of each layer. As shown in Algorithm \ref{alg_1}, all layers would connect to the first layer. When the nth layer is an odd-numbered layer, it would connect to all the previous even-numbered layers, when the nth layer is an even-numbered layer, it would connect to all the previous odd-numbered layers. The connection method of this algorithm can be reduced by about half the connections compared to the dense connection used by DenseNet.

For example, as shown in Fig. \ref{fig2}, when $n=8$, the $8th$ layer would only connect to the $1st$, $3rd$, $5th$ and $7th$ layers, and the connections to $2nd$, $4th$ and $6th$ layers are deleted. The connection method of Algorithm \ref{alg_1} is shown in the following formula:

\begin{equation}
x_{n}=\left\{\begin{array}{c}
H_{n}\left(\left[x_{1}, x_{2}, \cdots, x_{2^{i}}\right]\right), \text { if } n \% 2=1 \\
H_{n}\left(\left[x_{1}, x_{3}, \cdots, x_{2{ }^{i}+1}\right]\right), \text { if } n \% 2=0
\end{array}\right.
\label{eq2}
\end{equation}

\subsection{$ShortNet_2$}
The second new connection method we propose is more aggressive than the first one, and it deletes more connections. This connection method is similar to the harmonic dense connection used by HarDNet \cite{chao2019hardnet}, but the algorithm is not the same. As shown in Algorithm 2, when the layer is $n$, $x$ represents all values of $2^i-1$. $x$ is required to be less than $n$, and the $nth$ layer would connect to all $n-x$ layers.

\begin{algorithm}
\caption{$Shortnet_2$}
\label{alg_2}
\begin{algorithmic}
\REQUIRE {Stack 3 × 3 Convolution Layer in each block}
\ENSURE {$BN + ReLU + 3 \times 3 \ Conv$}
\FOR{layer $n$ is even}
\FOR{$i$ in $1$ to $5$}
\STATE {$x=2^i=1$}
\STATE{$y=n-x \ (x \leq n$)}
\STATE{layer $n$ connect to layer $y$}
\ENDFOR
\ENDFOR
\FOR{layer $n$ is odd}
\STATE{layer $n$ connect to layer $y$}
\ENDFOR
\end{algorithmic}
\end{algorithm}

For example, as shown in Fig. \ref{fig2}, when $n=8$, and $x=1,\ 3,\ 7$, then $n-x=1,\ 5,\ 7$. Therefore, the $8th$ layer would only connect to the $1st$, $5th$ and $7th$ layers. The model designed by connecting layers through this algorithm would be much smaller than the densely connected network model. The connection method of Algorithm \ref{alg_2} is shown in the following formula:

\begin{equation}
x_{n}=\left\{\begin{array}{c}
H_{n}\left(\left[x_{n-1}\right]\right), \text { if } n \% 2=1 \\
H_{n}\left(\left[x_{1}, \cdots, x_{n-2{ }^{i}+1}\right]\right), \text { if } n \% 2=0
\end{array}\right.
\label{eq3}
\end{equation}
	
\begin{table*}[]
\centering
\caption{Experimental Results}
\setlength{\tabcolsep}{3.5mm}{
\begin{tabular}{|c|c|c|c|c|c|c|c|c|c|}
\hline
\textbf{Model} & \textbf{\begin{tabular}[c]{@{}c@{}}C10 GPU\\ Time\\ (ms)\end{tabular}} & \textbf{\begin{tabular}[c]{@{}c@{}}C10\\ Error\\ (\%)\end{tabular}} & \textbf{\begin{tabular}[c]{@{}c@{}}SVHN GPU\\ Time\\ (ms)\end{tabular}} & \textbf{\begin{tabular}[c]{@{}c@{}}SVHN\\ Error\\ (\%)\end{tabular}} & \textbf{\begin{tabular}[c]{@{}c@{}}Flops\\ (G)\end{tabular}} & \textbf{\begin{tabular}[c]{@{}c@{}}MAdd\\ (G)\end{tabular}} & \textbf{\begin{tabular}[c]{@{}c@{}}Memory\\ (MB)\end{tabular}} & \textbf{\begin{tabular}[c]{@{}c@{}}\#Params\\ (M)\end{tabular}} & \textbf{\begin{tabular}[c]{@{}c@{}}MenR+W\\ (MB)\end{tabular}} \\ \hline
Baseline-43 & 72.83 & 14.00 & 72.64 & 5.95 & 509.38 & 1.02 & 6.08 & 2.17 & 25.93 \\ \hline
ShortNet\_1-43 & 61.17 & 13.59 & 58.97 & 5.65 & 374.00 & 0.75 & 4.60 & 1.59 & 18.92 \\ \hline
ShortNet\_2-43 & 52.48 & 14.09 & 50.61 & 5.48 & 256.44 & 0.51  & 4.00 & 0.97 & 13.74 \\ \hline
Baseline-53 & 94.25 & 13.38 & 92.11 & 5.92 & 783.20 & 1.56 & 7.37 & 3.15 & 35.46 \\ \hline
ShortNet\_1-53 & 71.19 & 13.36 & 69.57 & 5.63 & 536.76 & 1.07 & 5.41 & 2.16 & 24.56 \\ \hline
ShortNet\_2-53 & 58.14 & 14.08 & 55.34 & 6.59 & 334.76 & 0.67 & 4.37 & 1.20 & 16.05 \\ \hline
\end{tabular}
}
\label{tab2}
\end{table*}

\section{Experiment}
\subsection{Dataset}
The experiments use two small datasets (CIFAR-10 and SVHN) for image classification. CIFAR-10 \cite{krizhevsky2009learning} dataset consists of 60,000 images of 32 × 32 pixels for 10 categories, of which 50,000 are used for training and 10,000 are used for testing. The Street View House Numbers (SVHN) \cite{netzer2011reading} dataset is real house numbers captured by Google Street View. The entire dataset contains 73,257 training images and 26,032 testing images.

\subsection{Training}
For a fair comparison between $Baseline$ and $ShortNet_1$, $ShortNet_2$, we do not perform pre-training, model fine-tuning, and image augmentation.  In the process of training the model, we chose a common training hyperparameter, set the learning rate to 0.001, select Adam as the optimizer, and set the training batch size to 100 epochs. All network models run on the pytorch-1.10.0 framework on a single NVIDIA RTX 3050 4GB GPU.

\subsection{Testing}
TABLE \ref{tab2} shows the theoretical amount of floating point operation (Flops), the total number of network parameters (Params), the sum of memory read and memory write time (MemR+W), and multiplication and accumulations (MAdd) for 32 × 32 pixels images in dataset CIFAR-10. Compared with the dense connection, the connection method of Algorithm \ref{alg_1} deletes about half the number of connections, so the number of model parameters of $ShortNet_1$ is much smaller than that of $Baseline$. $ShortNet_1-53$ reduces the number of parameters from 3.15M  ($Baseline$) to 2.16M. In addition, the memory access of $ShortNet_1-53$ has also been optimized to be more suitable for platforms with limited memory in real life. From the test results of datasets CIFAR-10 and SVHN, it can be seen that the performance of the $ShortNet_1$ models with different depths is better than those of $Baseline$. For example, the test error rate of $ShortNet_1-43$ in CIFAR-10 is 13.59\%, which is lower than $Baseline-43$ of 14.00\%, while the inference time dropped from 72.83ms to 61.17ms. The calculation amount of the model can be shown by the inference time of the image. The calculation amount of $ShortNet_1$ is smaller than that of $Baseline$, and the performance of the network model is improved, which shows that the $ShortNet_1$ model is more efficient. In addition, the test results prove that for small datasets such as CIFAR-10 and SVHN, the dense connection is not the most suitable connection method. Reducing partial connections between layers often results in better performing network models.  Because for the input image of 32 × 32 pixels, an appropriate amount of connection plays the role of backpropagation and is enough to complete the feature extraction.

Compared with $ShortNet_1$, $ShortNet_2$ uses a more aggressive connection method algorithm, and deletes more connections, so the network model is smaller. In terms of the parameters of the model, the number of the parameters of $Baseline-53$ is 3.15M, while the number of the parameters of $ShortNet_2$ is only 1.20M, which is reduced by more than 60\% and greatly reduces the computational cost of the model. Compared with $Baseline$, the inference time of $ShortNet_2$  is much shorter. For a neural network with the depth of 43, the inference time of $ShortNet_2$ is 30\% faster than that of $Baseline$, and for the depth of 53, $ShortNet_2$ is 40\% faster. In the case of greatly shortened inference time, the test accuracy of the model does not reduce too much. The error rate of $ShortNet_2-43$ in SVHN is 5.48\% that is lower than that of $Baseline$ 5.95\%, and the error rate in CIFAR-10 is basically the same as $Baseline$. Because the increase of the depth of the model leads to the increase of the number of layers and the deletion of more connections, it is acceptable that the error rate of $ShortNet_2-53$ is slightly higher than that of $Baseline-53$.

DenseNet is limited by its high computational cost, and its application in different scenarios often cannot be the preferred neural network. The two new algorithms proposed in this article replace the dense connection between layers, and the improved DenseNet can have more applications in the future, such as TOOD \cite{feng2021tood} and YOLOF \cite{chen2021you} architectures for object detection scenes, SOLO \cite{wang2020solo} and DetectoRS \cite{qiao2021detectors} architectures for instance segmentation scenes, Panoptic FPN \cite{kirillov2019panoptic} and Mask2Former \cite{cheng2022masked} architectures for panoptic segmentation scenes. The improved DenseNet model is smaller and more suitable for application in Jetson Nano, mobile phones, FPGA and other platforms \cite{ju2022triplenet}. 

\section{Conclusion}
This paper proposes two algorithms for the connection between layers. Replacing the dense connections of DenseNet with these two algorithms can reduce the amount of model parameters and reduce the computational cost. In the case of evaluating network models in small datasets, reducing the connections between layers can effectively improve model performance and greatly speed up inference while ensuring accuracy. The new algorithms promote the improvement of DenseNet, making the improved DenseNet more suitable for application in real life.

\bibliographystyle{IEEEtran}
\bibliography{main}

\vspace{12pt}
\end{document}